\documentclass[12pt,draftcls,onecolumn]{IEEEtran}

\def \be {\begin{equation}}

\def \ee {\end{equation}}

\def \nn {\nonumber}

\usepackage[dvips]{graphicx}

\usepackage[cmex10]{amsmath}
\usepackage{amssymb}
\usepackage{color}

\begin{document}



\title{Adaptive Visual Tracking for Robotic Systems Without Image-Space Velocity Measurement}

\author{Hanlei Wang
\thanks{The author is with the Science and Technology on Space Intelligent Control Laboratory, Beijing Institute of
Control Engineering, Beijing 100190, China (e-mail: hlwang.bice@gmail.com).}}


\maketitle

\begin{abstract}
In this paper, we investigate the visual tracking problem for robotic systems without image-space velocity measurement, simultaneously taking into account the uncertainties of the camera model and the manipulator kinematics and dynamics. We propose a new image-space observer that exploits the image-space velocity information contained in the unknown kinematics, upon which, we design an adaptive controller without using the image-space velocity signal where the adaptations of the depth-rate-independent kinematic parameter and depth parameter are driven by both the image-space tracking errors and observation errors. The major superiority of the proposed observer-based adaptive controller lies in its simplicity and the separation of the handling of multiple uncertainties in visually servoed robotic systems, thus avoiding the overparametrization problem of the existing work. Using Lyapunov analysis, we demonstrate that the image-space tracking errors converge to zero asymptotically. The performance of the proposed adaptive control scheme is illustrated by a numerical simulation.


\end{abstract}



\section{Introduction}

It is generally believed that the incorporation of versatile sensory information (e.g., the information provided by joint position/velocity sensors, tip force/torque sensors, and vision systems) into the control system is an important aspect of intelligent robots. Mimicking the action of human beings, more and more manipulators are equipped with cameras to monitor their status and further to perform visual servoing so that the system can achieve certain robustness against model uncertainties (see, e.g., \cite{Espiau1992_TRA,Hutchinson1996_TRA}). Many results in the past years have been devoted to the visual servoing problem \cite{Espiau1992_TRA,Hashimoto1996_TRA,Hutchinson1996_TRA,Kelly2000_TMechtronics,Malis2002_TRA,Astolfi2002_TRA,Liu2006_TRO,Hu2009_TAC,Hu2010_JRNC,Mehta2012_MECH}. The visual servoing control schemes can in general be grouped into two classes (see, e.g., \cite{Hutchinson1996_TRA}). The first class (e.g., \cite{Espiau1992_TRA,Malis2002_TRA,Hu2010_JRNC}) is known as the position-based visual servoing, which simply takes the camera as a specific task-space sensor, i.e., the end-effector position/velocity information is obtained from the camera. One possible disadvantage of this scheme, as is frequently stated in the literature (e.g., \cite{Hutchinson1996_TRA,Liu2006_TRO}), is the requirement of the precise/extensive calibration. The second class (e.g., \cite{Kelly2000_TMechtronics,Astolfi2002_TRA,Liu2006_TRO}) is known as the image-based visual servoing, which directly utilizes the information of the concerned object in the image space and does not require the calibration of the camera. The advantage of the image-based visual servoing is now well known, i.e., the possible errors in establishing and calibrating the camera model are avoided.

As a standard control methodology, adaptive control has been shown to be adept at treating model uncertainties and be promising to achieve aggressive performance \cite{Slotine1991_Book}. Since the late 1980s,
numerous adaptive controllers for robot manipulators taking into account the nonlinear robot dynamics have been proposed (e.g., \cite{Craig1987_IJRR,Slotine1987_IJRR,Slotine1989_Aut}), and these controllers are all based on the linearity-in-parameters property of the manipulator dynamic model. The recent studies in \cite{Cheah2006_IJRR,Cheah2006_TAC,Braganza2005_CDC,Dixon2007_TAC} show how the linearity-in-parameters feature of the manipulator kinematics is exploited for performing adaptive tracking/regulation control in the case of existence of the kinematic uncertainties. An interesting
property of a visually servoed robotic system (with a fixed
camera) is that if the depth of the feature point with respect to the
camera frame is unknown but kept constant, the overall kinematics
of the system that describes the mapping from joint space to image space is linearly parameterized \cite{Cheah2006_IJRR}. This desirable feature of the overall kinematics, unfortunately, no longer holds in the case that the unknown depth is time varying since the depth acts as the denominator in the overall kinematics \cite{Liu2006_TRO,Cheah2007_ICRA,Wang2007_TRO,Cheah2010_Aut}. Via exploiting the respective linearity-in-parameters properties of the depth and the depth-independent interaction matrix, adaptive strategies are developed in \cite{Liu2006_TRO,Cheah2007_ICRA,Wang2007_TRO,Cheah2010_Aut,Li2012_ACC,Liang2014_ASJC} to handle the uncertain camera parameters. In particular, the adaptive visual tracking problem is resolved in \cite{Cheah2007_ICRA}, and the adaptive solutions to the visual regulation problem are given in \cite{Cheah2010_Aut,Liang2014_ASJC}, by designing appropriate control and adaptation laws to accommodate the uncertainties in the manipulator dynamics and kinematics and the camera model

However, one possible limitation of the above results which deal with the tracking problem is the requirement of image-space velocity measurement in the control input. One may notice that in the adaptive regulation algorithms given in \cite{Cheah2010_Aut,Liang2014_ASJC}, the control inputs do not need the image-space velocity measurement, yet the parameter adaptation laws do use the image-space velocity signal and in addition their extension to the more challenging tracking problem remains unclear. Also note
that if applying the approach in \cite{Braganza2005_CDC} to the visual
tracking problem with constant depth, the image-space velocity
can indeed be avoided in the kinematic parameter adaptation,
yet the control will still require the availability of the image-space
velocity. The image-space velocity is usually/commonly obtained by the standard numerical
differentiation of the image-space position information. It is well recognized that this velocity signal tends to be very noisy due in part to the relatively long processing time or delays of the image information, and thus it is undesirable to use image-space velocities in the control. One possible solution is given in \cite{Wang2010_TCST}, extending the result in \cite{Cheah2006_IJRR} to the case of time-varying uncertain depth. The limitation of \cite{Wang2010_TCST} lies in three aspects: 1) if we further accommodate the uncertain dynamics based on \cite{Wang2010_TCST}, the overparametrization and even nonlinear parametrization (due to the presence of the uncertain
depth in the denominator of an unknown term to be compensated
for) problems will occur (refer to  \cite[equation~(22)]{Wang2010_TCST}), and additionally the separation of the kinematic and dynamic uncertainties is impossible; 2) the determination of the controller parameters relies on some priori knowledge of the system model; 3) it requires high control activities to accommodate the variation
of the depth, due to the velocity-dependent feedback gain (which
means that the undesirable high-gain feedback is demanded in the
case that the manipulator motions at a high velocity). So, the best result that can be achieved by using the scheme in \cite{Wang2010_TCST} is still conservative. Other adaptive control schemes appear in \cite{Liu2006_Aut,Leite2011_IFAC,Wang2012_ICEICE}, where cascade-framework-based control schemes are proposed in \cite{Leite2011_IFAC,Wang2012_ICEICE}, and an observer-based controller is proposed in \cite{Liu2006_Aut},
which achieves the image-space trajectory tracking of electrically
driven robots with the desired armature current not involving
the image-space velocity. The results in \cite{Liu2006_Aut,Leite2011_IFAC,Wang2012_ICEICE}, in contrast to \cite{Wang2010_TCST}, take into consideration the uncertain robot kinematics and dynamics. Nevertheless, the results in \cite{Liu2006_Aut,Wang2012_ICEICE} can only deal with the case that the depth is constant, and the controller given in \cite{Leite2011_IFAC} needs to obtain the end-effector position with respect to the manipulator base frame so as to perform the kinematic parameter estimation (refer to \cite[equation (21)]{Leite2011_IFAC}) (which means that it is not a completely image-based visual servoing but a combination of image-based and position-based schemes, thus demanding the elaborate calibration and tending to be vulnerable to modeling errors). Moreover, the SDU factorization adopted in \cite{Leite2011_IFAC} (some detailed analysis appears in \cite{Lizarralde2013_Aut}) results in the complexity in both the controller design and stability analysis. Another limitation of \cite{Leite2011_IFAC} may be the requirement of the persistent excitation (PE) of the kinematic regressor (see the proof of Theorem 3 in \cite{Lizarralde2013_Aut})

In our opinion, the separation of the handling of multiple uncertainties of the system is highly preferred, whose superiority may be the avoidance of overparametrization, the simplification of the control scheme, and consequently better performance of the closed-loop system. Along this idea, in this paper, we propose an observer-based adaptive control scheme for visual tracking with time-varying depth (unlike the control schemes in \cite{Liu2006_Aut,Wang2012_ICEICE} that can only handle the constant depth case) and with uncertain manipulator kinematics and dynamics. The proposed adaptive controller avoids the measurement of image-space velocity and realizes the separation of the handling of three categories of parameter uncertainties. Using a depth-dependent quasi-Lyapunov function, we show the convergence of the image-space tracking errors. In contrast to the velocity-dependent-gain feedback and the overparametrization problem in \cite{Wang2010_TCST}, our control scheme employs a constant-gain feedback taking into account the uncertain manipulator dynamics and kinematics in addition to the uncertain camera model and achieves the separation of the handling of the depth, depth-rate-independent kinematic, and dynamic parameter uncertainties (avoiding the overparametrization or even the nonlinear parametrization). Moreover, the elaborate calibration and vulnerability to model uncertainties of \cite{Leite2011_IFAC} (due to the kinematic parameter estimation) are conquered by the proposed completely image-based servoing controller, and additionally, the PE condition associated with the kinematic regressor in \cite{Leite2011_IFAC} is not demanded in the proposed control scheme.

\section{Kinematics and Dynamics}

In this paper, we consider a visually servoed robotic system consisting of an $n$-DOF (degree-of-freedom) manipulator and a fixed pinhole uncalibrated camera (see, e.g., \cite{Forsyth2012_Book}), where the manipulator end-effector motion is mapped to the image space by the camera and it is assumed that the number of the feature points is $m$. The fact that the camera is not calibrated means that the extrinsic and intrinsic parameters of the camera are uncertain.

Let $x_i\in R^2$ (with the unit being pixel) represent the position of the projection of the $i$-th feature point on the image plane, and $r_i\in R^3$ denote the position of the $i$-th feature point with respect to the base frame of the manipulator, $i=1,\ldots,m$. Via the image Jacobian
matrix \cite{Hutchinson1996_TRA} or the interaction matrix \cite{Espiau1992_TRA}, the relationship
between the image-space velocity $\dot {x}_i$ and the feature-point velocity $\dot{r}_i$ can be written as \cite{Liu2006_TRO}
\be
\label{eq1} \dot {x}_i =\frac{1}{z_i(q)}\left(\bar D-x_i d_3^T\right) \dot r_i
\ee
where $z_i(q)\in R$ denotes the depth of the $i$-th feature point with respect to the camera frame, $\bar D\in R^{2\times 3}$ and $d_3\in R^3$ are taken from $D=\left[\bar D^T,d_3\right]^T$ which is the left $3\times 3$ portion of the perspective projection matrix, $N_i(x_i)=\bar D-x_i d_3^T\in R^{2\times 3}$ is called the depth-independent interaction matrix in \cite{Liu2006_TRO}, $i=1,\ldots,m$, and $q\in R^n$ denotes the joint position of the manipulator. In addition, it should be noted that {$z_i(q)=d_3^T r_i+d_0$ with $d_0$ being a constant} and $\dot z_i(q)=d_3^T \dot r_i$ (see also \cite{Liu2006_TRO}) and it is assumed that $z_i(q)$ is uniformly positive, $i=1,\dots,m$.

Equation (\ref{eq1}) can be rewritten as the following compact form
\be
\label{eq2}
\dot{x}=Z^{-1}(q) N(x)\dot{r}
\ee
where $x=\left[x_1^T, \dots, x_m^T\right]^T$, $r=\left[r_1^T, \dots, r_m^T\right]^T$, $Z(q)=\text{diag}\left[z_1(q)I_2,\dots,z_m(q)I_2\right]$ with $I_2$ being the $2\times2$ identity matrix, and $N(x)=\text{diag}\big[N_1(x_1), \dots,N_m(x_m)\big]$.

Let $v_0\in R^3$  denote the translational velocity of a reference point on the end-effector with respect to the manipulator base frame and $\omega_0\in R^3$ the angular velocity of the end-effector with respect to the manipulator base frame, which relate to the joint velocity $\dot q$ as \cite{Craig2005_Book,Spong2006_Book}
\be
\begin{bmatrix}
 v_0\\
\omega_0
\end{bmatrix}=J_r(q)\dot q
\ee
where $J_r(q)\in R^{6\times n}$ denotes the manipulator Jacobian matrix.

The relationship between the velocity of the $m$ feature points $\dot{r}$ and the manipulator joint velocity $\dot q$ can be written as \cite{Wang2007_TRO} (see also \cite{Hutchinson1996_TRA,Craig2005_Book,Spong2006_Book})
\be
\label{eq3}
\dot{r}=\underbrace{\begin{bmatrix}I_3 & -S(c_1)\\
\vdots & \vdots\\
I_3 & -S(c_m)\end{bmatrix}}_{J_f}J_r(q)\dot{q}
\ee
where $I_3$ is the $3\times 3$ identity matrix, $c_i\in R^3$ is the position vector of the $i$-th feature point with respect to the reference point on the manipulator end-effector, $i=1,\dots,m$, and the skew-symmetric form $S(b)$ is defined as $$S(b)=\begin{bmatrix}0 & -b_3 & b_2\\
b_3 &0& -b_1\\
-b_2& b_1 & 0\end{bmatrix}$$ for a 3-dimensional vector $b=\left[b_1,b_2,b_3\right]^T$.

The combination of (\ref{eq2}) and (\ref{eq3}) gives rise to the overall kinematic equation \cite{Cheah2007_ICRA,Wang2007_TRO,Cheah2010_Aut}, i.e.,
\be
\label{eq4}
\dot{x}=Z^{-1}(q)\underbrace{N(x)J_f J_r(q)}_{J(q,x)}\dot{q}
\ee
where $J(q,x)$ is a Jacobian matrix that does not depend on the depth (also referred to as the depth-independent image Jacobian matrix in \cite{Wang2007_TRO}). The exploitation of the structure of (\ref{eq1}) allows $J(q,x)$ to be decomposed as
\begin{equation}
\label{eqa1}
J(q,x)=\underbrace{\left(I_m\otimes \bar D\right)J_fJ_r(q)}_{J_z^\perp(q)}-X \underbrace{(I_m\otimes d_3^T)J_f J_r(q)}_{J_z(q)}
\end{equation}
where $I_m$ is the $m\times m$ identity matrix, the matrix $X=\text{diag}\left[x_i,i=1,\dots,m\right]$, $\otimes$ denotes the Kronecker product \cite{Brewer1978_TCS}, $J_z^\perp(q)$ is a Jacobian matrix that maps the joint velocity $\dot q$ to a plane which is parallel to the image plane, and $J_z(q)$ is a Jacobian matrix that describes the relationship between the changing rate of the depth vector $z(q)=\left[z_1(q),\dots,z_m(q)\right]^T$ and $\dot{q}$ (see, e.g., \cite{Liu2006_TRO}), i.e.,
\be
\label{eqa2}
\dot{z}(q)=J_z(q)\dot{q}.
\ee
It is worth remarking that the existence of the second term on the right side of (\ref{eqa1}) is due to the variation of the depth vector $z(q)$ while that of the first one is independent of the variation of
$z(q)$. Therefore, $J_z^\perp(q)$ is called the \emph{depth-rate-independent Jacobian matrix}.

We now make the following assumption.

\emph{Assumption 1:} The number of the manipulator DOFs and that of the feature points satisfy the constraint that $n\ge 2m$  and $m\le 3$, and the three feature points are non-collinear in the case $m=3$. Furthermore, for $\forall u=\left[u_1^T,\dots,u_m^T\right]^T$ with $u_i\in R^2$, $i=1,\dots,m$, the rank of $N(u)J_f$ is $2m$.

\emph{Remark 1\footnote{The discussions on the cases of $m=2$ and $m=3$ are largely due to the constructive comments from one anonymous reviewer.}:} From \cite[Proposition 1]{Liu2006_TRO}, we obtain that $\text{rank}\left[N_i(u_i)\right]=2$, $\forall i$. Next, we discuss the rank of $N(u)J_f$ for $m=1$, $m=2$, and $m=3$, respectively.

1) In the case $m=1$, it is straightforward to obtain that $J_f$ has full row rank and thus $\text{rank}\left[N(u)J_f\right]=2$ (see also \cite{Liu2006_TRO,Wang2007_TRO}).

2) In the case $m=2$, the rank of $N(u)J_f$ is equal to that of the matrix
$J_f^T N^T(u)=\begin{bmatrix}
I_3 & I_3 \\
S(c_1) & S(c_2)
\end{bmatrix}\begin{bmatrix}
N_1^T(u_1) & 0_{3\times 2}\\
0_{3\times 2} & N_2^T(u_2)
\end{bmatrix}.$
Now consider the following linear equation with $\mu_1,\mu_2\in R^3$ being the unknowns
\be
\label{eq:a1}
\begin{bmatrix}
I_3 & I_3 \\
S(c_1) & S(c_2)
\end{bmatrix}\begin{bmatrix}\mu_1\\ \mu_2\end{bmatrix}=0.
\ee
As is well known, the rank of the skew-symmetric matrix $S(b)$ is $2$ for $\forall b\ne 0$, and therefore the rank of $S(c_2 - c_1)$ is
2, which leads us to obtain from the standard matrix theory
that the rank of the coefficient matrix $J_f$ is 5. According to the standard theory of linear equations, the solutions of equation (\ref{eq:a1}) constitute a one-dimensional space with the elements being of the form $\left[\mu_1^T, \mu_2^T\right]^T=k\left[c_1^T-c_2^T, c_2^T-c_1^T\right]^T$ where $k$ is an arbitrary constant. Let us now consider the following linear equation with $\lambda_i$, $i=1,\dots,4$ being the unknowns
\be
\label{eq:a2}
\begin{bmatrix}
N_1^T(u_1) & 0_{3\times 2}\\
0_{3\times 2} & N_2^T(u_2)
\end{bmatrix}\begin{bmatrix}\lambda_1 \\ \vdots \\ \lambda_4\end{bmatrix}=k\begin{bmatrix}c_1-c_2\\ c_2-c_1\end{bmatrix}.
\ee
If $c_1-c_2$ is not in the intersection of the range spaces of $N_1^T(u_1)$ and $N_2^T(u_2)$, equation (\ref{eq:a2}) has a solution only in the case that $k=0$, and this solution is $\lambda_i=0$, $i=1,\dots,4$. Hence, the rank of $N(u)J_f$ is $4$.

3) In the case $m=3$, 
from the standard matrix theory, the rank of $J_f^T$ is equal to that of the following matrix (which is obtained by the elementary row operation of $J_f^T$)
$$\begin{bmatrix}
I_3 & I_3 & I_3\\
0_{3\times3} &S(c_2-c_1) & S(c_3-c_1)
\end{bmatrix}.$$
To determine the rank of this matrix, we have to identify that of $B=\begin{bmatrix}S(c_2-c_1) & S(c_3-c_1)\end{bmatrix}$. Suppose that there is a nonzero vector $\mu\in R^3$ such that
$B^T \mu=0$,
which then means that $\mu$ is parallel to $c_2-c_1$ and $c_3-c_1$ simultaneously.
 Obviously, this will not happen since the three feature points are non-collinear. Therefore, the rank of $B$ is 3 and consequently the rank of $J_f^T$ is 6. Then, we
obtain from the standard theory of linear equations that the null space of $J_f^T$ is a set containing three independent basis vectors, whose elements can be expressed as $k_1\left[c_1^T-c_2^T,c_2^T-c_1^T,0_3^T\right]^T+k_2\left[0_3^T,c_2^T-c_3^T,c_3^T-c_2^T\right]^T+k_3\left[c_1^T-c_3^T,0_3^T,c_3^T-c_1^T\right]^T$ with $k_1$, $k_2$, and $k_3$ being arbitrary constants. Now consider the following linear equation with $\lambda_i$, $i=1,\dots,6$ being the unknowns
\begin{align}
\label{eq:a3}
&\begin{bmatrix}
N_1^T(u_1) & 0_{3\times 2} & 0_{3\times2}\\
0_{3\times 2} & N_2^T(u_2) & 0_{3\times2}\\
0_{3\times 2} & 0_{3\times 2} & N_3^T(u_3)
\end{bmatrix}\begin{bmatrix}\lambda_1 \\ \vdots \\ \lambda_6\end{bmatrix}\nn\\
&=k_1\begin{bmatrix}c_1-c_2\\c_2-c_1\\0_3\end{bmatrix}+k_2\begin{bmatrix}0_3\\c_2-c_3\\c_3-c_2\end{bmatrix}+k_3\begin{bmatrix}c_1-c_3\\0_3\\c_3-c_1\end{bmatrix}.
\end{align}
If none of the nonzero elements in $\text{span}\left\{c_1-c_2,c_1-c_3\right\}$ are in the range space of $N_1^T(u_1)$, none of the nonzero elements in $\text{span}\left\{c_2-c_3,c_2-c_1\right\}$ are in the range space of $N_2^T(u_2)$, and none of the nonzero elements in $\text{span}\left\{c_3-c_1,c_3-c_2\right\}$ are in the range space of $N_3^T(u_3)$, equation (\ref{eq:a3}) has only one solution $\lambda_i=0$, $i=1,\dots,6$. Hence, the rank of $J_f^T N^T (u)$ in this case is $6$.

\emph{Remark 2:} The rank of $N(u)J_f$ has been discussed in \cite[p.~616]{Wang2007_TRO}. Yet, the analysis there is neither complete nor rigorous for the cases $m=2$ and $m=3$. Here, it is demonstrated that $N(u)J_f$ has full row rank if the relative position vectors between the feature points in the manipulator base frame satisfy certain conditions. The proof of the fact that $\text{rank}(J_f)=5$ for the case $m=2$ and that $\text{rank}(J_f)=6$ for the case $m=3$ has already been given in \cite[p.~616]{Wang2007_TRO}, yet a different approach is used here to prove this fact. For more complete and detailed discussions as well as the vivid explanations of the singularity issues associated with the case of three feature points (i.e., $m=3$), please refer to \cite{Michel1993_Report}.

We further make the following assumption to facilitate the controller design and stability analysis in the sequel.

\emph{Assumption 2:} For $\forall u=\left[u_1^T,\dots,u_m^T\right]^T$ with $u_i\in R^2$, $i=1,\dots,m$, the matrix $J(q,u)=N(u)J_f J_r(q)$ has full row rank in the case that Assumption 1 holds.

Assumption 2 holds if the manipulator is away from the singular configuration and the manipulator end-effector and the camera are in the nonsingular relative configuration. In fact, from Assumption 1, we know that $\text{rank}[N(u)J_f]=2m$.  Since the manipulator is assumed to be away from the singular configuration, we obtain $\text{rank}\left[J_r(q)\right]=\min\left\{n,6\right\}\ge 2m$. From \cite[p.~210]{Meyer2000_Book}, the rank of $J(q,u)$ can be determined as
$
\text{rank}\left[J(q,u)\right]=\text{rank}\left[J_r(q)\right]-\text{dim}\left[N^\ast(N(u)J_f)\cap R^\ast(J_r(q))\right]
$,
 where $N^\ast(N(u)J_f)$ denotes the null space of $N(u)J_f$ and $R^\ast(J_r(q))$ the range space of $J_r(q)$. The vectors in the range space of $J_r(q)$ that denote the velocities of the feature points motioning towards the pinhole of the camera, obviously, lie in the null space of $N(u)J_f$ since, physically, the image-space velocities corresponding to these vectors are zero. The assumption that the end-effector and the camera are in the nonsingular relative configuration ensures that the rank of $J(q,u)$ is the largest, i.e., only $\min\left\{n,6\right\}-2m$ basis vectors in $R^\ast(J_r(q))$ lie in the null space of $N(u)J_f$. Then, we obtain $\text{rank}\left[J(q,u)\right]=\min\left\{n,6\right\}-(\min\left\{n,6\right\}-2m)=2m$. In the special case that $n\ge 6$, from \cite[p.~220]{Meyer2000_Book}, we have $\text{rank}\left[J(q,u)\right]=\text{rank}\left[N(u)J_f\right]=2m$, which implies that the nonsingular relative configuration is always ensured for $n\ge 6$.



The overall kinematics (\ref{eq4}) has the following property.

\textit{Property 1:} The two quantities $Z(q)\psi$ and $\dot{Z}(q)\phi$ can be linearly parameterized \cite{Liu2006_TRO,Cheah2007_ICRA}, i.e.,
\begin{align}
Z(q)\psi=&Y_z(q,\psi)a_z\\
\dot{Z}(q)\phi=&\bar{Y}_z(q,\dot{q},\phi)a_z
\end{align}
where $\psi=\left[\psi_1^T,\dots,\psi_m^T\right]^T$ and $\phi=\left[\phi_1^T,\dots,\phi_m^T\right]^T$ with $\psi_i\in R^2$ and $\phi_i\in R^2$, $i=1,\dots,m$, which also directly yields
\be
\label{eqa3}
\Phi J_z(q)\dot q=\dot Z(q)\phi=\bar{Y}_z(q,\dot q,\phi)a_z
\ee
where $\Phi=\text{diag}\left[\phi_i,i=1,\dots,m\right]$, $a_z\in R^{p_1}$ is the unknown depth parameter vector, and $Y_z(q,\psi)\in R^{(2m)\times p_1}$ and $\bar{Y}_z(q,\dot{q},\phi)\in R^{(2m)\times p_1}$ are two regressor matrices.
In addition, $J(q,x)\dot{q}$ can also be linearly parameterized \cite{Cheah2007_ICRA}, which gives
\be
\label{eqa4}
J_z^\perp(q)\dot q=Y_z^\perp(q,\dot{q})a_z^\perp
\ee
where $a_z^\perp\in R^{p_2}$ is the unknown depth-rate-independent kinematic parameter vector, and $Y_z^\perp(q,\dot{q})\in R^{(2m)\times p_2}$ is the depth-rate-independent kinematic regressor matrix. Therefore, $J(q,x)\dot q$ can be parameterized as [by (\ref{eqa3}) and (\ref{eqa4})]
\be
\label{eqa5}
J(q,x)\dot q=Y_z^\perp(q,\dot q)a_z^\perp-\bar{Y}_z(q,\dot q,x)a_z.
\ee



The equations of motion of the manipulator can be written as \cite{Slotine1991_Book,Spong2006_Book}
\be
\label{eq5}
M(q)\ddot{q}+C(q,\dot{q})\dot{q}+g(q)=\tau
\ee
where $M(q)\in R^{n\times n}$ is the inertia matrix, $C(q,\dot{q})\in R^{n\times n}$ is the Coriolis and centrifugal matrix, $g(q)\in R^n$ is the gravitational torque, and $\tau\in R^n$ is the exerted joint torque. Three fundamental properties associated with the dynamics (\ref{eq5}) that shall be useful for the subsequent controller design and stability analysis are listed as follows (see, e.g., \cite{Slotine1991_Book,Spong2006_Book}).

\textit{Property 2:} The inertia matrix $M (q )$ is symmetric and uniformly positive definite.

\textit{Property 3:} The Coriolis and centrifugal matrix $C(q,\dot{q})$ can be suitably selected such
that $\dot {M}(q) - 2C(q,\dot{q})$ is skew-symmetric.

\textit{Property 4:} The dynamics (\ref{eq5}) depends linearly on an unknown constant dynamic parameter vector $a_d\in R^{p_3}$, and thus
\begin{equation}
\label{eq7}
M \left( {q } \right)\dot {\xi } + C \left( {q ,\dot {q} }
\right)\xi + g \left( {q } \right) = Y_d \left( {q ,\dot {q} ,\xi
,\dot {\xi }} \right)a_d
\end{equation}
where $Y_d\left( q ,\dot {q} ,\xi
,\dot {\xi } \right)\in R^{n\times p_3}$ is the
dynamic regressor matrix, $\xi \in R^n$ is a differentiable vector, and
$\dot {\xi }$ is the derivative of the vector $\xi$ with respect to time.

\section{Observer-Based Adaptive Tracking Control}

In this section, we investigate the adaptive visual tracking for robotic systems with time-varying depth and with uncertain kinematics and dynamics. We will at first develop an image-space observer, and then, based on this observer, we propose an adaptive tracking controller without involving image-space velocity measurement to realize the asymptotic trajectory tracking in the image space, i.e., $x-x_d\to 0$ and $\dot{x}-\dot{x}_d\to0$ as $t\to\infty$, where $x_d$ denotes the desired trajectory in the image space and we assume that $x_d$, $\dot{x}_d$, and $\ddot{x}_d$ are all bounded.

The image-space observer is designed as
\begin{align}
\label{eq10}
\dot{x}_o=&\hat{Z}^{-1}(q)\hat{J}(q,x)\dot{q}-\frac{1}{2}\hat{Z}^{-1}(q)\hat{\dot{Z}}(q)\nn\\
&\times\left(x_o-x_d\right)-\alpha\left(x_o-x\right)
\end{align}
where $x_o$ denotes the observed quantity of the image-space position, $\alpha$ is a positive design constant, $\hat{Z}(q)$ and $\hat{\dot{Z}}(q)$ are the estimates of $Z(q)$ and $\dot{Z}(q)$, respectively, which are obtained by replacing $a_z$ in $Z(q),\dot{Z}(q)$ with its estimate $\hat{a}_z$, and $\hat J(q,x)$ is the estimate of $J(q,x)$, which is obtained by replacing $a_z^\perp$ and $a_z$ in $J(q,x)$ with their estimates $\hat{a}_z^\perp$ and $\hat a_z$, respectively. The employment of the second term on the right side of (\ref{eq10}) is to accommodate the variation of the depth.

The closed-loop observer dynamics can be written as
\begin{align}
\label{eq11}
\Delta\dot{x}_o=&\hat{Z}^{-1}(q)\hat{J}(q,x)\dot{q}-Z^{-1}(q)J(q)\dot{q}-\frac{1}{2}\hat{Z}^{-1}(q)\hat{\dot{Z}}(q)\nn\\
&\times\left(x_o-x_d\right)-\alpha\Delta x_o
\end{align}
where $\Delta x_o=x_o-x$ is the image-space observation error. Equation (\ref{eq11}) can be further formulated as
\begin{align}
\label{eq12}
Z(q)\Delta \dot{x}_o=&\left[Z(q)-\hat{Z}(q)\right]\hat{Z}^{-1}(q)\hat{J}(q,x)\dot{q}\nn\\
&+\hat{J}(q,x)\dot{q}-J(q,x)\dot{q}\nn\\
&-\frac{1}{2}Z(q)\hat{Z}^{-1}(q)\hat{\dot{Z}}(q)\left(x_o- x_d\right)\nn\\&
-\alpha Z(q)\Delta x_o.
\end{align}
Let us rewrite (\ref{eq12}) as (by {Property 1})
\begin{align}
\label{eq13}
Z&(q)\Delta \dot{x}_o+\frac{1}{2}\dot{Z}(q)(x_o-x_d)\nn\\
=&-Y_z\left(q,\hat{Z}^{-1}(q)\hat{J}(q,x)\dot{q}\right)\Delta a_z\nn\\
&+Y_z^\perp(q,\dot{q})\Delta a_z^\perp-\bar Y_z\left(q,\dot q,x\right)\Delta a_z\nn\\
&\underbrace{+\frac{1}{2}\dot{Z}(q)\left(x_o-x_d\right)-\frac{1}{2}Z(q)\hat{Z}^{-1}(q)\hat{\dot{Z}}(q)\left(x_o-x_d\right)}_{\Pi}\nn\\
&-\alpha Z(q)\Delta x_o
\end{align}
where $\Delta a_z=\hat{a}_z-a_z$ and $\Delta a_z^\perp=\hat{a}_z^\perp-a_z^\perp$ are the depth and depth-rate-independent kinematic parameter estimation errors, respectively, and the term $\Pi$ can be interestingly written as (again by {Property 1})
\begin{align}
\Pi=&\frac{1}{2}\left[\dot{Z}(q)-\hat{\dot{Z}}(q)\right]\left(x_o- x_d\right)\nn\\
&+\frac{1}{2}\left[\hat{Z}(q)-Z(q)\right]\hat{Z}^{-1}(q)\hat{\dot{Z}}(q)\left(x_o-x_d\right)\nn\\
=&-\frac{1}{2}\bar{Y}_z(q,\dot{q},x_o-x_d)\Delta a_z\nn\\
&+\frac{1}{2}Y_z\left(q,\hat{Z}^{-1}(q)\hat{\dot{Z}}(q)\left(x_o-x_d\right)\right)\Delta a_z.
\end{align}
In this way, equation (\ref{eq13}) can be rewritten as
\begin{align}
\label{eq15}
Z&(q)\Delta \dot{x}_o+\frac{1}{2}\dot{Z}(q)(x_o-x_d)\nn\\
=&-\alpha Z(q)\Delta x_o+Y_z^\perp(q,\dot{q})\Delta a_z^\perp-Y_z^\ast\Delta a_z
\end{align}
where the combined depth regressor $Y_z^\ast$ is defined by
\begin{align}
Y_z^\ast=&Y_z\left(q,\hat{Z}^{-1}(q)\hat{J}(q,x)\dot{q}\right)+\bar{Y}_z\left(q,\dot{q},x+\frac{x_o-x_d}{2}\right)\nn\\
&-\frac{1}{2}Y_z\left(q,\hat{Z}^{-1}(q)\hat{\dot{Z}}(q)\left(x_o-x_d\right)\right).
\end{align}

Next, we develop an adaptive controller based on the observed quantities generated by the observer (\ref{eq10}), and the kinematic equation (\ref{eqa2}) and the decomposition property of $J(q,x)$ given by equation (\ref{eqa1}) will be exploited for the adaptive controller design.

Let us define a joint reference velocity as
\be
\label{eq18}
\dot{q}_r=\big[\underbrace{\hat{J}\left(q,(x_o+x_d)/2\right)}_{\hat{J}^\ast}\big]^{+}\left[\hat Z(q)\dot{x}_r\right]
\ee
where $\hat{J}^{\ast+}=\hat{J}^{\ast T}(\hat{J}^\ast\hat{J}^{\ast T})^{-1}$ is the standard generalized inverse of the modified estimated Jacobian matrix $\hat{J}^\ast$ [which is obtained by replacing $a_z^\perp$ and $a_z$ in $J\left(q,(x_o+x_d)/2\right)$ with $\hat a_z^\perp$ and $\hat a_z$, respectively], and $\dot x_r=\dot{x}_d-\gamma(x_o-x_d)$ with $\gamma$ being a positive design constant. Differentiating (\ref{eq18}) with respect to time gives the joint reference acceleration
\begin{align}
\label{eq19}
\ddot{q}_r=&\hat{J}^{\ast +}\left[\hat{Z}(q)\ddot{x}_r+\dot{\hat{Z}}(q)\dot{x}_r-\dot{\hat{J}}^\ast \dot{q}_r\right]\nn\\
&+(I_n-\hat J^{\ast+}\hat J^\ast)\dot{\hat J}^{\ast T} \hat J^{\ast+ T}\dot q_r
\end{align}
where the standard result concerning the time derivative of $\hat J^{\ast +}$ is used and $I_n$ is the $n\times n$ identity matrix.
As can be clearly seen from (\ref{eq19}), the variable $\ddot{q}_r$  does not involve the measurement of the image-space velocity $\dot{x}$.

\emph{Remark 3:} 
The use of the modified estimated Jacobian matrix $\hat J^\ast$ instead of the estimated Jacobian matrix $\hat{J}(q,x)$ is to accommodate the effect of the time-varying depth and to avoid the image-space velocity measurement in deriving the joint reference acceleration.

Then, define a joint-space sliding vector
\be
\label{eq20}
s=\dot{q}-\dot{q}_r.
\ee
Using $\hat{J}^{\ast}$ to premultiply both sides of (\ref{eq20}) and exploiting {Property 1} gives
\begin{align}
\hat{J}^{\ast} s=&\hat{J}(q,x)\dot{q}+\frac{1}{2}\hat{\dot{Z}}(q)(-x_o-x_d+2x)-\hat{Z}(q)\dot{x}_r\nn\\
=&Z(q)\left[\dot{x}-\dot{x}_d+\gamma(x_o-x_d)\right]\nn\\&+\frac{1}{2}\dot{Z}(q)(\Delta x-\Delta x_o)
+Y_z^\perp(q,\dot{q})\Delta a_z^\perp\nn\\&-\Big[\underbrace{\frac{1}{2}\bar{Y}_z(q,\dot{q},x_o+x_d)+Y_z(q,\dot{x}_r)}_{Y_z^{\ast\ast}}\Big]\Delta a_z
\end{align}
where $\Delta x=x-x_d$ is the image-space position tracking error.

Now we propose the control law as
\be
\label{eq22}
\tau=-\hat{J}^{\ast T} K \hat{J}^\ast s +Y_d(q,\dot{q},\dot{q}_r,\ddot{q}_r)\hat{a}_d
\ee
where $K$ is a symmetric positive definite matrix and $\hat{a}_d$ is the estimate of $a_d$. The adaptation laws for the estimated parameters $\hat a_d$, $\hat{a}_z^\perp$, and $\hat{a}_z$ are given as
\begin{align}
\label{eq23}
\dot{\hat{a}}_d=&-\Gamma_d Y_d^T(q,\dot{q},\dot{q}_r,\ddot{q}_r)s\\
\label{eq24}
\dot{\hat{a}}_z^\perp=&\Gamma_z^\perp Y_z^{\perp T}(q,\dot{q})\left(\Delta x-\Delta x_o\right)\\
\label{eq25}
\dot{\hat{a}}_z=&-\Gamma_z \left(Y_z^{\ast\ast T}\Delta x-Y_z^{\ast T}\Delta x_o\right)
\end{align}
where $\Gamma_d$, $\Gamma_z^\perp$, and $\Gamma_z$ are all symmetric positive definite matrices.

Substituting the control law (\ref{eq22}) into the manipulator dynamics (\ref{eq5}) yields
\be
\label{eq26}
M(q)\dot{s}+C(q,\dot{q})s=-\hat{J}^{\ast T}K\hat{J}^{\ast}s+Y_d(q,\dot{q},\dot{q}_r,\ddot{q}_r)\Delta a_d
\ee
where $\Delta a_d=\hat{a}_d-a_d$ is the dynamic parameter estimation error.

The closed-loop behavior of the system can then be described by
\be
\label{eq27}
\begin{cases}
Z(q)\Delta \dot{x}_o+(1/2)\dot{Z}(q)(\Delta x_o+\Delta x)\\
=-\alpha Z(q)\Delta x_o+Y_z^\perp(q,\dot{q})\Delta a_z^\perp-Y_z^\ast\Delta a_z,\\
Z(q)\Delta \dot{x}+(1/2)\dot{Z}(q)(\Delta x-\Delta x_o)\\
=-\gamma Z(q)(x_o-x_d)
-Y_z^\perp(q,\dot{q})\Delta a_z^\perp+Y_z^{\ast\ast}\Delta a_z+\hat{J}^{\ast}s,\\
M(q)\dot{s}+C(q,\dot{q})s\\
=-\hat{J}^{\ast T}K\hat{J}^{\ast}s+Y_d(q,\dot{q},\dot{q}_r,\ddot{q}_r)\Delta a_d
\end{cases}
\ee
and the parameter adaptation laws (\ref{eq23}), (\ref{eq24}), and (\ref{eq25}).

We are presently ready to formulate the following theorem.

\emph{Theorem 1:} The observer (\ref{eq10}), the control (\ref{eq22}), and the adaptation laws (\ref{eq23}), (\ref{eq24}), (\ref{eq25}) for the visually servoed robotic system (\ref{eq4}), (\ref{eq5}) guarantee the convergence of the image-space tracking errors if $\alpha>\gamma/3$, i.e., $\Delta x\to 0$ and $\Delta \dot x\to 0$ as $t\to\infty$.

\emph{Proof:} Following \cite{Slotine1987_IJRR,Ortega1989_Aut}, we consider the Lyapunov-like function candidate $
V_1=(1/2)s^T M(q)s+(1/2)\Delta a_d^T \Gamma_d^{-1} \Delta a_d
$, whose time derivative along the trajectories of the third subsystem of (\ref{eq27}) and (\ref{eq23}) can be written as
$
\dot{V}_1=-s^T \hat{J}^{\ast T} K \hat{J}^{\ast}s\le 0
$ (exploiting {Property 3}),
which implies that $s\in \cal L_\infty$, $\hat{J}^{\ast}s\in {\cal L}_2$, and $\hat{a}_d\in \cal L_\infty$. The fact that $\hat{J}^{\ast}s\in {\cal L}_2$ and $Z(q)$ is uniformly positive definite yields the result that $\int_0^t s^T \hat{J}^{\ast T}Z^{-1}(q)\hat{J}^{\ast}s dr\le l_M,\forall t\ge 0$ for some positive constant $l_M$.

Let us consider the following depth-dependent nonnegative function
\begin{align}
\label{eq30}
V_2=&\frac{1}{2}\Delta x_o^T Z(q)\Delta x_o+\frac{1}{2}\Delta x^T Z(q)\Delta x\nn\\
&+\frac{1}{2}\Delta a_z^{\perp T} \Gamma_z^{\perp -1 }\Delta a_z^\perp+\frac{1}{2}\Delta a_z^T \Gamma_z^{-1}\Delta a_z\nn\\
&+\underbrace{\frac{1}{\gamma}\left[l_M-\int_0^t s^T \hat{J}^{\ast T}Z^{-1}(q)\hat{J}^{\ast}s dr\right]}_{\Pi^\ast}
\end{align}
where the employment of the term ${\Pi}^\ast$ follows the typical practice (see, e.g., \cite[p. 118]{Lozano2000_Book}).

The time derivative of $V_2$ along the trajectories of the upper two subsystems of (\ref{eq27}) can be written as
\begin{align}
\label{eq31}
\dot{V}_2=&-\alpha \Delta x_o^T Z(q) \Delta x_o-\gamma \Delta x^T Z(q)(x_o-x_d)\nn\\
&-(\Delta x-\Delta x_o)^T Y_z^\perp(q,\dot{q})\Delta a_z^\perp\nn\\
&+\left(\Delta x^T Y_z^{\ast\ast}-\Delta x_o^T Y_z^\ast\right)\Delta a_z\nn\\
&+\Delta a_z^{\perp T} \Gamma_z^{\perp -1}\dot{\hat a}_z^\perp+\Delta a_z^T \Gamma_z^{-1}\dot{\hat{a}}_z\nn\\
&+\Delta x^T \hat{J}^{\ast}s-\frac{1}{\gamma} s^T \hat{J}^{\ast T} Z^{-1}(q)\hat{J}^{\ast}s.
\end{align}
Substituting the adaptation laws (\ref{eq24}) and (\ref{eq25}) into (\ref{eq31}) gives
\begin{align}
\label{eq32}
\dot{V}_2=&-\alpha \Delta x_o^T Z(q)\Delta x_o-\gamma \Delta x^T Z(q)\Delta x\nn\\
&-\gamma \Delta x^T Z(q)\Delta x_o+\Delta x^T \hat{J}^{\ast}(q)s\nn\\
&-\frac{1}{\gamma} s^T \hat{J}^{\ast T} Z^{-1}(q)\hat{J}^{\ast}s.
\end{align}
Using the following result obtained from the standard theory of inequalities
\begin{align*}
\Delta x^T \hat{J}^{\ast}s&\le\frac{1}{4}\gamma \Delta x^T Z(q) \Delta x+\frac{1}{\gamma} s^T \hat{J}^{\ast T} Z^{-1}(q)\hat{J}^{\ast}s
\end{align*}
we obtain from (\ref{eq32}) that
\begin{align}
\label{eq33}
\dot{V}_2\le&-\alpha\Delta x_o^T Z(q)\Delta x_o-\gamma \Delta x^T Z(q)\Delta x_o-\frac{3\gamma}{4}\Delta x^T Z(q)\Delta x\nn\\
=&-\begin{bmatrix}\Delta x_o\\
\Delta x\end{bmatrix}^T\underbrace{\begin{bmatrix}\alpha Z(q) & (\gamma/2)Z(q)\\
(\gamma/2)Z(q) & (3\gamma/4)Z(q)\end{bmatrix}}_{H}\begin{bmatrix}\Delta x_o\\
\Delta x\end{bmatrix}\le 0
\end{align}
since the matrix $H$ is uniformly positive definite under the condition $\alpha>\gamma/3$, according to the standard matrix theory. The inequality (\ref{eq33}) as well as the definition of $V_2$ given by (\ref{eq30}) yields the result that $\Delta x_o\in {\cal L}_2\cap \cal L_\infty$, $\Delta x\in {\cal L}_2\cap \cal L_\infty$, $\hat{a}_z^\perp\in \cal L_\infty$, and $\hat{a}_z\in \cal L_\infty$. If $\text{rank}(\hat{J}^\ast)=2m$, we obtain from the standard matrix theory that $\hat J^{\ast +}$ is bounded. Then, we obtain that $\dot{q}_r\in \cal L_\infty$ from equation (\ref{eq18}) since $\hat{Z}(q)$ is bounded and $\dot{x}_r\in {\cal L}_\infty$. From the result that $s\in {\cal L}_\infty$, we have that $\dot{q}\in \cal L_\infty$. From (\ref{eq10}), we have that $\dot{x}_o\in \cal L_\infty$, which further gives rise to the result that $\ddot{x}_r\in \cal L_\infty$. From the adaptation laws (\ref{eq24}) and (\ref{eq25}), we have that $\dot{\hat a}_z^\perp\in \cal L_\infty$ and $\dot{\hat{a}}_z\in \cal L_\infty$, which mean that $\dot{\hat{Z}}(q)$ and $\dot{\hat{J}}^\ast$ are bounded. Therefore, we obtain that $\ddot{q}_r\in \cal L_\infty$ from (\ref{eq19}). From (\ref{eq26}), we obtain that $\dot{s}\in \cal L_\infty$ since $M(q)$ is uniformly positive definite (by Property 2), which, plus the result that $\ddot{q}_r\in \cal L_\infty$, yields the conclusion that $\ddot{q}\in \cal L_\infty$. Then, from the kinematics (\ref{eq4}) and its differentiation with respect to time,
we obtain that $\dot{x}\in \cal L_\infty$ and $\ddot{x}\in \cal L_\infty$. We also obtain that $\ddot{x}_o\in \cal L_\infty$ from the differentiation of equation (\ref{eq10}). Then, we have that $\Delta \dot x_o\in {\cal L}_\infty$, $\Delta \dot x\in{\cal L}_\infty$, $\Delta \ddot x_o\in {\cal L}_\infty$, and $\Delta \ddot x\in{\cal L}_\infty$. Hence, $\Delta x_o$, $\Delta x$, $\Delta \dot{x}_o$, and $\Delta \dot{x}$ are all uniformly continuous. From the properties of square-integrable and uniformly continuous functions \cite[p. 117]{Lozano2000_Book}, we obtain that $\Delta x_o\to 0$ and $\Delta x\to 0$ as $t\to\infty$. Then, from Barbalat's Lemma \cite{Slotine1991_Book}, we have that $\Delta \dot{x}_o\to 0$ and $\Delta \dot{x}\to 0$ as $t\to\infty$.  \hfill{\small $\blacksquare$}


\emph{Remark 4:} The avoidance of image-space velocity measurement is achieved at the kinematic level, which results in the separation of the handling of the kinematic and dynamic uncertainties. In addition, the cascaded feature of the closed-loop system facilitates the stability analysis.

\emph{Remark 5:}

\begin{enumerate}
\item  Compared with the results in \cite{Cheah2006_IJRR,Cheah2007_ICRA,Wang2010_TCST,Leite2011_IFAC}, the novel points of our result mainly lie in the proposed observer (\ref{eq10}), the definition of the reference velocity (\ref{eq18}), the image-space-velocity-free adaptation law (\ref{eq25}), and the proposed depth-dependent quasi-Lyapunov function (\ref{eq30}) as well as the associated stability analysis. The adaptation law (\ref{eq24}) for updating $\hat{a}_z^\perp$ coincides with the one in \cite{Liu2006_Aut}\footnote{The task-space observer and the desired armature current given in \cite{Liu2006_Aut} (which deals with the adaptive control of electrically driven robots) make us believe that one can obtain the solution for rigid robots (a reduced case of electrically driven robots) from \cite{Liu2006_Aut} and will find that the adaptation law (\ref{eq24}) is in essence the same as this solution.}, \cite{Wang2012_ICEICE}, yet the results in \cite{Liu2006_Aut,Wang2012_ICEICE} are confined to the simpler case of constant depth.  The control law (\ref{eq22}) as well as the dynamic parameter adaptation law (\ref{eq23}) is basically the same as the one in \cite{Cheah2006_IJRR} (i.e., an extension of \cite{Slotine1987_IJRR} to handle both the uncertain kinematics and dynamics), yet employ a new estimated Jacobian matrix $\hat J^\ast$ and new reference velocity and acceleration.

    \item The simplicity of the proposed control scheme is reflected in the aspects that the overparametrization when accommodating the uncertain dynamics is avoided and the constant-gain feedback is adopted (unlike the result in, e.g., \cite{Wang2010_TCST}), and that the explicit measurement of the feature-point position with respect to the manipulator base frame is not required (in contrast with \cite{Leite2011_IFAC}).

\end{enumerate}

\emph{Remark 6:} The standard projection approach \cite{Ioannou1996_Book} can be applied to the adaptation laws (\ref{eq24}) and (\ref{eq25}) so that $\hat J^\ast$ has full row rank [this originates from the fact that $J\left(q,(x_o+x_d)/2\right)$ has full row rank according to {Assumption 2}] and $\hat{Z}(q)$ is uniformly positive definite during the adaptation process (see also \cite{Cheah2007_ICRA,Cheah2010_Aut}).

\section{Simulation Results}

In this section, we present the simulation results to show the performance of the proposed observer-based adaptive controller. We consider a visually servoed robotic system that includes a typical three-DOF manipulator and a fixed camera, as is shown in Fig. 1, and the number of the feature points that are under consideration is set as one. The focal length of the camera is set as $f=0.15 \text{ m}$ and the two scaling factors of the camera are set to be the same value $\beta=900.0$. The three axes of the camera frame (denoted by $X_C$, $Y_C$ and $Z_C$, respectively) are assumed to be aligned with the axes $Y_0$, $Z_0$, and $X_0$ of the manipulator base frame, respectively, yet there is an offset $d_C=5.0\text{ m}$ along the axis $Z_C$ between the origins of the two frames. The lengths of the three links of the manipulator are $l_1=2.0 \text{ m}$, $l_2=2.0 \text{ m}$, and $l_3=2.0 \text{ m}$. The mass and inertia properties of the manipulator are not listed due to the space limitation. The sampling period is chosen to be 5 ms.

The controller parameters are determined as $K=0.001I_2$, $\alpha=10.0$, $\gamma=10.0$, $\Gamma_d=300.0I_8$, $\Gamma_z^\perp=600.0I_2$ and $\Gamma_z=0.2I_3$. The initial estimates of the kinematic parameters (including the camera parameters) are chosen as $\hat{l}_2(0)=\hat{l}_3(0)=3.0\text{ m}$, $\hat{d}_C(0)=3.0\text{ m}$, $\hat{f}(0)=0.1 \text{ m}$, and $\hat{\beta}(0)=700.0$. The initial estimate of the dynamic parameter vector is chosen as $\hat{a}_d(0)=\left[0_6^T, 15, 0\right]^T$. The desired trajectory in the image space is given as
$
x_d=\left[
45+20\cos(\pi t /3),
65+20\sin(\pi t /3)\right]^T$.
The simulation results are shown in Fig. 2  and Fig. 3. From Fig. 2, we see that the image-space position tracking errors indeed converge to zero asymptotically. Fig. 3 gives the responses of the actual and estimated depths during the motion of the manipulator. It seems that the estimated depth tends to approach the actual depth. Although the convergence of the depth estimation error does not occur, the
asymptotic image-space trajectory tracking is still realized.


%
%
%

\begin{figure}
\centering
\begin{minipage}[t]{1.0\linewidth}
\centering
\includegraphics[width=3.0in]{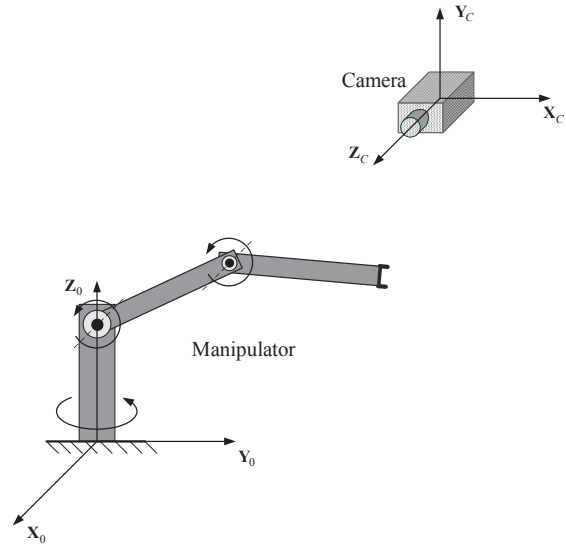}
\caption{Three-DOF manipulator with a fixed camera}\label{fig:side:a}
\end{minipage}%
\end{figure}

\begin{figure}
\centering
\begin{minipage}[t]{1.0\linewidth}
\centering
\includegraphics[width=3.0in]{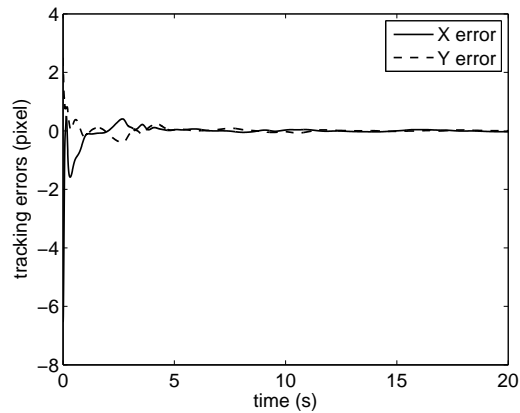}
\caption{Image-space position tracking errors}\label{fig:side:a}
\end{minipage}%
\end{figure}


\begin{figure}
\centering
\begin{minipage}[t]{1.0\linewidth}
\centering
\includegraphics[width=3.0in]{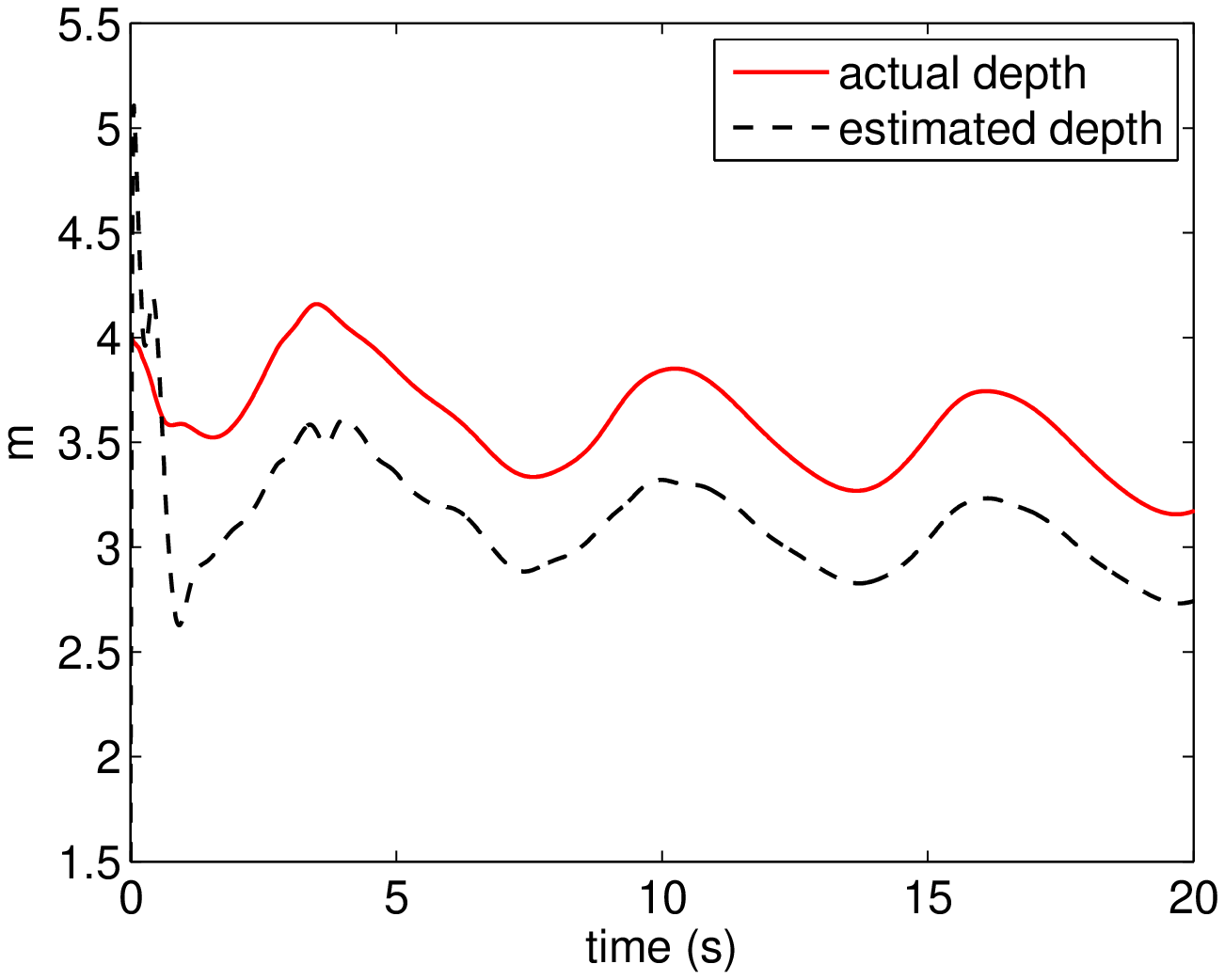}
\caption{Actual and estimated depths}\label{fig:side:a}
\end{minipage}%
\end{figure}

%
%


\section{Conclusion}

In this paper, we have examined the visual tracking problem for robotic systems with uncertain camera model and uncertain manipulator kinematics and dynamics, and the image-space velocity is assumed to be unavailable. To achieve visual tracking without image-space velocity measurement, we propose a novel image-space observer and an adaptive controller based on the observed quantities, which yield a cascade closed-loop robotic system. Using a depth-dependent quasi-Lyapunov function plus the standard Lyapunov-like function for analyzing the Slotine and Li adaptive controller, we demonstrate that the image-space tracking errors converge to zero. We also show the asymptotic convergence of the image-space observation errors. A simulation is conducted to show the performance of the proposed observer-based adaptive controller.

\section*{Acknowledgment}

The author would like to thank the anonymous reviewers and the Associate Editor of Automatica for their helpful comments on the paper.

\bibliographystyle{IEEETran}        
\bibliography{..//Reference_list_Wang}           
\end{document}